%% file: root.tex
\documentclass[letterpaper, 10 pt, journal, twoside]{ieeetran}
\usepackage{amsmath,amsfonts}
\usepackage{algorithmic}
\usepackage{array}
\usepackage[caption=false,font=normalsize,labelfont=sf,textfont=sf]{subfig}
\usepackage{textcomp}
\usepackage{stfloats}
\usepackage{url}
\usepackage{verbatim}
\usepackage{graphicx}
\def\BibTeX{{\rm B\kern-.05em{\sc i\kern-.025em b}\kern-.08em
    T\kern-.1667em\lower.7ex\hbox{E}\kern-.125emX}}
\usepackage{balance}
\usepackage{preamble}

\usepackage{pifont}
\usepackage{fontawesome}
\usepackage{pgfplotstable}
\usepackage{hyperref}
\hypersetup{
    colorlinks=true,
    citecolor=black,
    filecolor=black,
    urlcolor=gray,
    breaklinks=true,
}
\usepackage{soul}
\setstcolor{red}

\begin{document}
\title{Stereo Hand-Object Reconstruction for Human-to-Robot Handover}
\author{Yik Lung Pang, Alessio Xompero, Changjae Oh, Andrea Cavallaro
\thanks{Manuscript received: December, 6, 2024; Revised March, 3, 2025; Accepted March, 30, 2025.}
\thanks{This paper was recommended for publication by Editor Abhinav Valada upon evaluation of the Associate Editor and Reviewers' comments.
This work was supported in part by the CHIST-ERA program through the project CORSMAL, under UK EPSRC grant EP/S031715/1 and in part by the Royal Society Research Grant RGS\textbackslash R2\textbackslash242051. This project also made use of time on Tier 2 HPC facility JADE2, funded by EPSRC (EP/T022205/1).}
\thanks{Yik Lung Pang, Alessio Xompero, and Changjae Oh are with Centre for Intelligent Sensing, Queen Mary University of London, UK
        {\tt\footnotesize \{y.l.pang,a.xompero,c.oh\}@qmul.ac.uk}}%
\thanks{Andrea Cavallaro is with Idiap Research Institute and École Polytechnique Fédérale de Lausanne, Switzerland
        {\tt\footnotesize a.cavallaro@idiap.ch}}%
\thanks{Digital Object Identifier (DOI): see top of this page.}}

\markboth{IEEE Robotics and Automation Letters. Preprint Version. Accepted March, 2025}
{Pang \MakeLowercase{\textit{et al.}}: Stereo Hand-Object Reconstruction for Human-to-Robot Handover} 

\maketitle

\begin{abstract}
Jointly estimating hand and object shape facilitates the grasping task in human-to-robot handovers. Relying on hand-crafted prior knowledge about the geometric structure of the object fails when generalising to unseen objects, and depth sensors fail to detect transparent objects such as drinking glasses. In this work, we propose a method for hand-object reconstruction that combines single-view reconstructions probabilistically to form a coherent stereo reconstruction. We learn 3D shape priors from a large synthetic hand-object dataset, and use RGB inputs to better capture transparent objects. We show that our method reduces the object Chamfer distance compared to existing RGB based hand-object reconstruction methods on single view and stereo settings. We process the reconstructed hand-object shape with a projection-based outlier removal step and use the output to guide a human-to-robot handover pipeline with wide-baseline stereo RGB cameras. Our hand-object reconstruction enables a robot to successfully receive a diverse range of household objects from the human.
\end{abstract}

\begin{IEEEkeywords}
Human-robot collaboration, physical human-robot interaction, deep learning for visual perception, perception for grasping and manipulation
\end{IEEEkeywords}

\section{Introduction}
\IEEEPARstart{A}{s} humans and robots share the same workspace~\cite{kumar2020survey}, exchanging objects is an essential skill to enable smooth collaboration. 
We consider the scenario where a human hands various everyday objects over to a robot, which aims to grasp the object while avoiding the human (human-to-robot handover~\cite{ortenzi2021object,rosenberger2020object,sanchez2020benchmark}) and delivers the object at a predefined location (see Fig.~\ref{fig:robot_setup}). Objects can vary in their physical properties, such as appearance, transparency, shape, size, and mass~\cite{xompero2022corsmal}. Container-like objects, such as cups, drinking glasses, or boxes, can be empty or filled with content, affecting the perception of the appearance or the weight of the object~\cite{sanchez2020benchmark,pang2021towards}. 
Reconstructing the hand and object shapes is essential in this highly interactive scenario. The object shape informs the robot on how to grasp the object and the hand shape allows the robot to avoid dangerous contact with the human.

Visual modalities (RGB and depth images) are used in hand-object shape reconstruction for human-to-robot handover. Previous works~\cite{yang2020human,rosenberger2020object,yang2021reactive,yang2022model,wang2022goal} used depth sensors to capture the pointcloud of the scene and applied segmentation models to extract the pointcloud of the hand and object. However, transparent objects are difficult to detect with depth sensors~\cite{sajjan2020clear} and object shapes are incomplete due to partial visibility or occlusions when using a single camera~\cite{rosenberger2020object}. Therefore, depth completion methods are devised to fill in the missing depth values~\cite{yu2024depth}, but the pointcloud is only filled in on the visible side of the object and a depth camera is still required.
As RGB-only cameras are more commonly available and less expensive than RGB-D cameras, some works~\cite{sanchez2020benchmark,pang2021towards} used wide-baseline stereo RGB cameras to address partial visibility and occlusions of the object. However, these methods are limited to only container-like objects (rotationally symmetric shape) and assume that the object is held upright during the handover due to the presence of a content.
 
\input{robot_setup}

To address the above challenges, in this paper we propose StereoHO\footnote{Code and videos of the experiments are available at\\ \url{https://qm-ipalab.github.io/StereoHO/}}, a hand-object reconstruction method for wide-baseline stereo inputs. Our method quantifies the uncertainty of single-view predictions, e.g. caused by self-occlusions, by predicting low-dimensional embeddings of hand and object shapes as a probability distribution over learned codebooks of discrete shape embeddings. Individual predictions from each camera views are aggregated and decoded jointly to form a coherent stereo reconstruction. We train our model on synthetic 3D hand-object data~\cite{hasson19_obman} with a large amount of objects and without specific assumptions about the object's geometry to ensure its generalisation ability. We integrate StereoHO in the first human-to-robot handover pipeline that jointly reconstructs hand and object shapes from stereo RGB inputs while tracking the hand over time. The pipeline  also ensures multi-view consistency and reconstruction quality for accurate robot grasp predictions. Experiments show that StereoHO outperforms previous methods in single-view and stereo settings for object reconstruction, while remaining competitive for hand reconstruction in terms of Chamfer distance. Furthermore, our pipeline can successfully account for objects of diverse shapes and appearances, including filled/empty transparent containers, boxes and thin objects.

\section{Related work}

Previous works~\cite{rosenberger2020object,yang2021reactive,yang2022model,wang2022goal,sanchez2020benchmark,pang2021towards,yang2020human} in human-to-robot handover focused on marker-less settings that are more applicable to daily human-to-robot interactions.
Human grasps were classified into different types to select the corresponding predefined safe robot grasp direction and enable the robot to receive the object~\cite{yang2020human}. However, this approach does not consider diverse object shapes and limits the grasp types that the robot can handle.
Capturing the shape of the hand and the object improves the perception, and using available grasp estimation methods allows to predict possible grasps on the object shape~\cite{rosenberger2020object,yang2021reactive,yang2022model,wang2022goal,sanchez2020benchmark,pang2021towards}. Grasps that would cause a collision between the robot and the human are removed for safety. A two-stage system was proposed to perform human-to-robot handovers of everyday objects~\cite{rosenberger2020object}. The first stage segments the person and the object from the input RGB image. The depth image is lifted into a pointcloud and the background is removed using the segmentation masks. The second stage adds an imaginary plane in the grasping direction to approximate the task as a top-down grasp estimation problem using GG-CNN~\cite{morrison2018closing}. Using 6-DoF grasps allows more freedom in the robot's movement, enabling more natural handovers~\cite{yang2021reactive}.

\begin{figure}[t!]
\subfloat[]{\includegraphics[width=0.43\columnwidth,trim={0 11cm 18cm 0},clip]{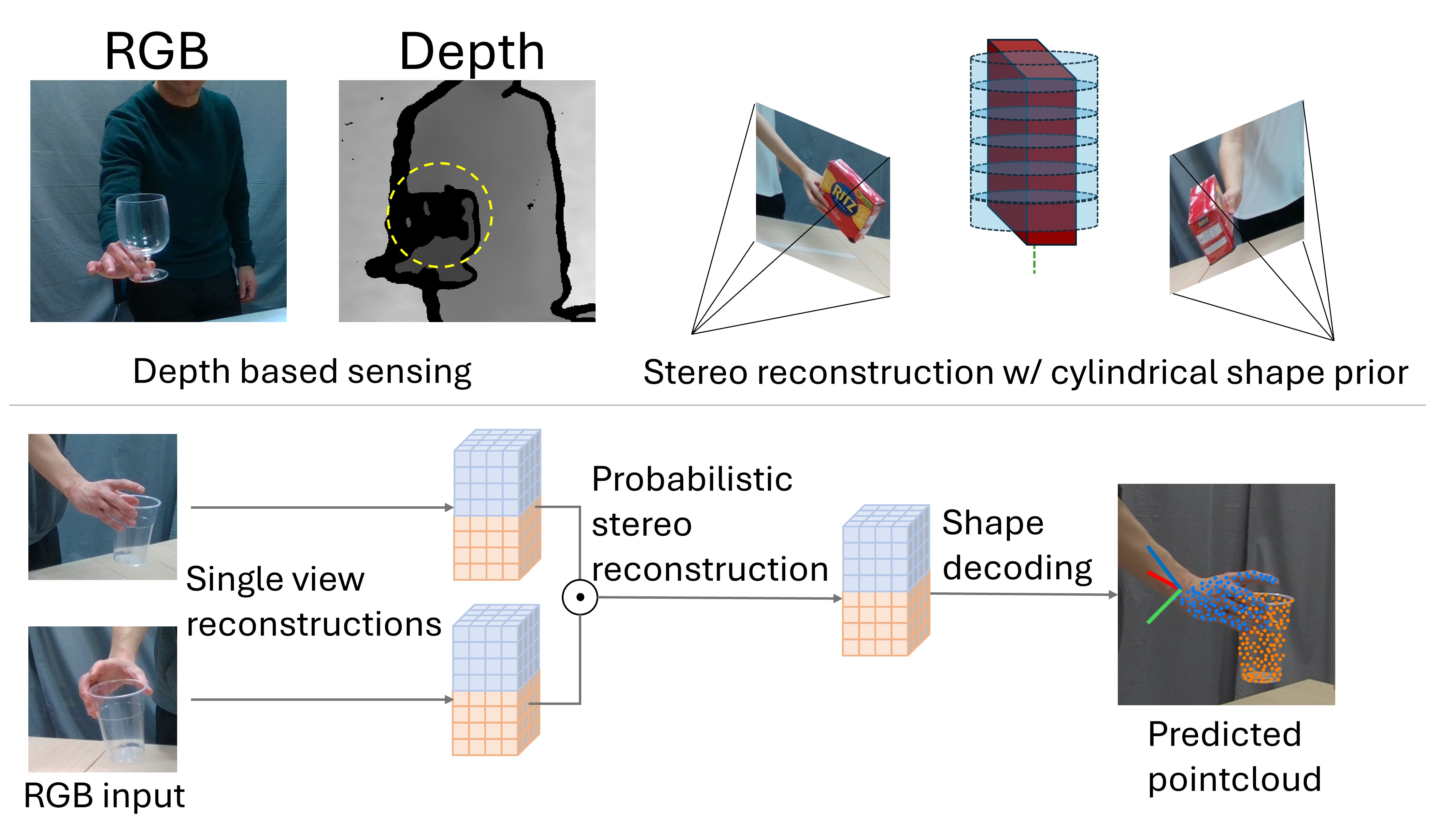}}
\subfloat[]{\includegraphics[width=0.55\columnwidth,trim={15cm 11cm 0cm 0},clip]{comparison.jpg}}
\caption{
    Limitations of existing human-to-robot handover approaches. 
    (a) Depth-based sensing fails on transparent objects. (b) Relying on shape priors does not generalise.
    }
\label{fig:comparison}
\end{figure}

Methods for human-to-robot handovers~\cite{rosenberger2020object,yang2021reactive,yang2022model,wang2022goal} and grasp estimation~\cite{morrison2018closing,mousavian20196}, rely on pointcloud as a representation for hand and object shape. 
However, pointclouds of objects are view-dependent and are incomplete in the presence of occlusions~\cite{rosenberger2020object} or when the objects are transparent and reflective~\cite{sajjan2020clear}. Incomplete object shapes or missing pointclouds for transparent objects (see Fig.~\ref{fig:comparison}) can results in inaccurate grasp estimations.
To overcome this issue, depth restoration is a method to complete the pointcloud of transparent objects~\cite{sajjan2020clear}.
However, a depth camera is required which is not as commonly available as RGB cameras. Moreover, the completed pointcloud often only consists of the visible side of the object~\cite{yu2024depth}. An alternative representation for object shapes is Signed Distance Dields (SDFs) that encode into a vector the signed distance between a set of points in 3D and their closest point on the 3D object~\cite{chen2019learning}. As an implicit representation of smooth and continuous surfaces, SDFs distinguish the inside and
outside parts of an arbitrary object and are easier to learn compared to pointclouds~\cite{chen2019learning}.
None of the existing works has used SDFs for reconstructing object-only or hand-object shapes for more accurate human-to-robot handovers.

Other methods use stereo RGB images as input to reconstruct the shape of the object~\cite{sanchez2020benchmark,pang2021towards}. The shape of containers, such as drinking glasses and cups, was reconstructed from stereo RGB images by assuming a rotationally symmetric geometry (see Fig.~\ref{fig:comparison}) and exploiting the predicted segmentation masks of the object~\cite{xompero2020multi}. This enables human-to-robot handover by grasping at the centroid of the object~\cite{sanchez2020benchmark}. Human safety is considered by estimating the 3D hand keypoints~\cite{pang2021towards}. Robot grasps are sampled along the vertical axis of the object and grasp points that are too close to the hand keypoints are removed. However, these methods do not estimate 6-DoF grasps and thus limit the naturalness of the robot movement as the object must maintain an upright orientation during the handover. The assumption on the object shape is easily invalidated by non-container objects, causing failure in the object shape estimation. 

Joint hand-object reconstruction from RGB images in an interaction scenario is challenging due to severe mutual occlusions between the hand and the object~\cite{hasson19_obman}. Large hand-object interaction datasets, both synthetic and collected in real environments~\cite{hasson19_obman,chao2021dexycb,yang2022oakink}, have helped improve the reconstruction quality. Methods for hand-object reconstruction can be grouped into single-view and multi-view, and multi-view methods can be further split into dense and sparse. Single-view reconstruction methods~\cite{hasson19_obman,ye2022s,chen2023gsdf,choi2024handnerf} learn 3D shape priors from hand-object datasets. However, completing the shape of the unobserved parts of an object remains challenging. Multi-view methods~\cite{hampali2022hand,swamy2023showme,qu2023novel,ye2023diffusion,pang2024sparse} overcome the problems of occlusion and unobserved object parts by increasing the number of input views. Dense multi-view methods~\cite{hampali2022hand,swamy2023showme,qu2023novel,ye2023diffusion} use a large number of input views  to produce high quality reconstructions but are  computationally expensive, limiting the applicability in reactive scenarios. Sparse multi-view methods~\cite{pang2024sparse} balances the reconstruction quality of dense methods with the speed of single-view methods. However, training only on synthetic data can affect the model's ability to generalise to real world scenarios (sim-to-real gap)~\cite{pang2024sparse}.

\input{overview_recon}

\section{Proposed method}

In this section, we introduce StereoHO for stereo hand-object reconstruction and present our human-to-robot handover pipeline based on StereoHO. 

\subsection{Stereo-based RGB hand-object reconstruction}

We design StereoHO to predict the joint surface of the hand ($H$) and object ($O$), represented as pointcloud (\mbox{$\mathcal{P}=[\mathcal{P}_H, \mathcal{P}_O]$}), from the RGB images of a stereo-based camera. Images are cropped and centred on the person's hand holding an object. We formulate stereo-based hand-object shape reconstruction as a view-dependant image-to-shape classification task (encoding) followed by an aggregation step~\cite{ye2022s} and a transformation of the aggregated shape codes into the final joint hand-object pointcloud (decoding). Fig.~\ref{fig:overview_recon} illustrates the inference pipeline of StereoHO. To avoid erroneously including background clutter in the reconstructed shape, we  include the segmentation masks of the object ($M_O$) and the hand ($M_H$) as input to the model. We also predict the wrist pose ($T_H$) for each view to relate the 3D shapes with the cameras views, assuming the camera poses are not available.

For the \textit{encoding} part, we define a 3D grid centred on the hand wrist $T_{Hv}$, where $v \in \{L, R\}$ is the camera view. For each view $v$, we project the 3D position of each voxel into the image space using the wrist pose and associate each voxel with an image embedding, extracted with a convolutional neural network (e.g., ResNet-18~\cite{he2016deep}), at the corresponding pixel location~\cite{ye2022s}. We use 3D convolutional layers to transform the resulting 4D tensor into a probability distribution $P_v \in [0,1]^{D\times D \times D \times C}$ with respect to a codebook of $C$ shape embeddings, $\mathcal{C} = \{e_c \in \mathbb{R}^S\}_{c=1}^C$.

We combine the predicted probability distributions from each view into a single coherent prediction via element-wise multiplication: $P = P_L P_R$. We thus define another 4D tensor of shape embeddings that are selected from $\mathcal{C}$ based on the index of the highest probability for each element of the grid, $E \in \mathbb{R}^{D \times D \times D \times S}$. The same procedure is performed independently for the hand and the object, resulting in the two shape-based tensors, $E_O$ and $E_H$, respectively.

For the \textit{decoding} part, we apply hand and object specialised networks, consisting of 3D upconvolutional layers, to obtain the pointclouds $\mathcal{P}_H$ and $\mathcal{P}_O$ from $E_O$ and $E_H$, respectively. Each network (SDF decoder) predict SDF values, $\hat{s} \in \mathbb{R}^{D \times D\times D}$. We then sample the Truncated-SDF (T-SDF) at discrete positions and only the values within a threshold of 1~cm to the surface are kept as the predicted pointclouds. We ensure multi-view consistency by re-projecting the predicted hand and object pointcloud $\mathcal{P}_H$ and $\mathcal{P}_O$ in the images of both camera views using the wrist poses $T_{Hv}$, and by filtering out any outlier $\mathcal{P}_B$ with respect to the segmentation masks $M_H$ and $M_O$, resulting in the final prediction $\mathcal{P}'$. 

Next, we detail how we learn the codebooks of 3D shape embeddings and the neural network to reconstruct the hand-object pointcloud from the shape embeddings, and the image-to-shape embeddings. As hand-object data is hard to obtain and annotate, we train the components of StereoHO only on synthetic data. Note that the use of segmentation masks facilitate sim-to-real transfer as they are domain-invariant. 

\input{overview_robot}

\subsection{Learning discrete 3D shape embeddings}

We learn two discrete codebooks, one for the objects $\mathcal{C}_O$ and one for the hands $\mathcal{C}_H$, by training a 3D Patch-wise Encoding Variational Autoencoder with Vector Quantization~\cite{autosdf2022} on a synthetic hand-object dataset~\cite{hasson19_obman}. For each shape (object or hand) in the training set, we transform the shape representation from a 3D mesh to SDFs, discretized in a 3D grid (T-SDFs), $s \in \mathbb{R}^{D \times D\times D}$. The model encodes local patches of the T-SDF with 3D convolutional layers into a shape code $z_e \in \mathbb{R}^S$, maps each shape code to the closest quantized shape embedding $e \in \mathbb{R}^S$ in the codebook (vector quantization~\cite{van2017neural}), and reconstructs the shape $\hat{s}$. Through mapping the continuous embeddings into a finite set of learned embeddings, vector quantization reduces the high-dimensionality of the shape embeddings~\cite{van2017neural}. The SDF decoder then retrieves the T-SDF values $\hat{s}$ from $e$.

We use the following loss to train each model by simultaneously learning the autoencoder\footnote{We tested the two autoencoders on ObMan~\cite{hasson19_obman}, achieving a Chamfer distance of 0.015~cm\textsuperscript{2} and 0.437~cm\textsuperscript{2} for the hand and object, respectively.} parameters and codebook, 
\begin{equation}
    \mathcal{L}_{ae} = |s-\hat{s}| + ||sg[z_e]-e||^2_2 + \beta||z_e-sg[e]||^2_2,
\end{equation}
where $sg[\cdot]$ is the stop gradient operator. 
The hyper-parameter $\beta$ controls the scale of the commitment loss (last term), that helps stabilize the training~\cite{autosdf2022}. We initialise the codebook with a uniform distribution. After training the models for object and hand, we only keep the trained SDF decoders and codebooks and link them to the image-to-shape encoder.

\subsection{Training of the image-to-shape encoder}

We train an image-to-shape encoder that transforms the input image and segmentation masks into a probability distributions over the hand and object codebook. We use ResNet-18~\cite{he2016deep} pre-trained on ImageNet\cite{deng2009imagenet} as our image encoder and 3D convolutional layers as prediction heads for the hand and object.
We train all layers with a weighted cross-entropy loss function defined as
\begin{equation}
    \mathcal{L}_{ce} = -\sum_{c}^{C}{w_c}\cdot p_{gt}(c) \cdot \text{log } p(c),
\end{equation}
where $p_{gt}$ and $w_c$ are the ground-truth probability and weight for index $c$. We set a lower weight for the index $c$ corresponding to empty space, as the majority of the  surroundings of the hand and object in the 3D grid is empty.

\subsection{Handover with stereo hand-object reconstruction}

We integrate StereoHO into a pipeline for human-to-robot handovers (see Fig.~\ref{fig:overview_robot}) and assume that i) the handover task starts with the person already holding the object and ii) two cameras are placed on each side of the robot, maximising the visible area of the hand and object.

For each frame of the \textit{approaching phase}, StereoHO reconstructs the pointclouds of the hand and the object using image crops, pose of the wrist, and segmentation masks from each view. Therefore, for each camera view and for each frame, a hand-object detector~\cite{shan2020understanding} estimates the bounding boxes of both the hand and the object, $b_{H}, b_{O} \in \mathbb{R}^4$. We use Fast Segment Anything~\cite{zhao2023fast} to segment the object ($M_O$) within the object bounding box~\cite{zhao2023fast} and we crop the image around the hand using $b_{H}$. From the image crop, we use FrankMocap~\cite{rong2021frankmocap} to predict the pose of the wrist ($T_{H}$) and the segmentation mask of the hand ($M_H$) after extracting the silhouette from the hand shape. We use both segmentation masks to remove the background from the image crop. We triangulate the centroid of the object mask and verify that the reprojection error in each view is less than 5 pixels. Note that the pipeline does not perform the handover (failure case) if, within 10~s, no hand-object pair is detected in either of the views or the triangulation is not validated.
Since the reconstruction quality can vary across frames, significantly affecting the estimated robot grasp, we compute the Intersection over Union (IoU) of the convex hull of the pointclouds with the segmentation masks, respectively. We update the pointclouds if there is an improvement in reconstruction quality (i.e., IoU $>$ IoU$^\ast$).

To enable the robot to receive the object, we use 6-DoF GraspNet~\cite{mousavian20196} to estimate a set of $N$ candidate grasps, $\mathcal{G} = \{g_n \in SE(3)\}_{n=1}^N$, on the object pointcloud $\mathcal{P}'_O$.
To ensure human safety, we construct a 3D bounding box of the size of the gripper around each grasp, and we filter out any grasps with points from the hand point cloud $\mathcal{P}'_H$ in its bounding box. Using the predicted wrist pose in the robot coordinate frame, $T_{HB}=T_{WB}T_{HW}$, where $T_{WB}$ is the robot base transform, we transform the predicted grasps from the hand-wrist to the robot coordinate frame. For each timestep, the robot control selects the grasp $g$ closest to the current gripper 6-DoF pose. We define a standoff position 15~cm away from the grasp point in the direction of the grasp. Once the robot reaches the standoff position, the robot enters a close-loop \textit{grasping stage} and closes its fingers when the gripper reaches the selected grasp. 
After grasping, the robot retracts to the standoff position, delivers the object at a predefined location on the table, and returns to its initial position.

\subsection{Parameter settings}

We set the T-SDF dimension to $D=128$. The codebook size is $C=512$ with embedding size $S=128$. For $\mathcal{L}_{ae}$, we set $\beta=1.0$. For $\mathcal{L}_{ce}$, we set $w_c=0.25$ for the index $c$ corresponding to empty space and $w_c=0.75$ for the rest of the codebook. The robot base transform $T_{WB}$ is obtained using the hand-eye calibration process~\cite{tsai1989new}. We estimate $N=200$ grasp candidates for each handover.

\section{Validation}

\subsection{Datasets}

We train StereoHO on the ObMan~\cite{hasson19_obman} dataset and test it on DexYCB~\cite{chao2021dexycb}. The training set of ObMan contains 87,190 images of synthetic handheld objects rendered on top of real images. Objects were selected from ShapeNet~\cite{chang2015shapenet} based on 8 categories (Fig.~\ref{fig:objects_corsmal}), such as bottles, bowls, cans, jars, knifes, cellphones, cameras and remote controls, and grasped by the MANO hand model~\cite{romero2022embodied}. DexYCB is a multi-view dataset with videos of a person manipulating one of 20 everyday objects selected from YCB-Video~\cite{xiang2018posecnn}.

\subsection{Handover setup}

\begin{figure}[t!]
    \centering

    \includegraphics[width=0.98\columnwidth]{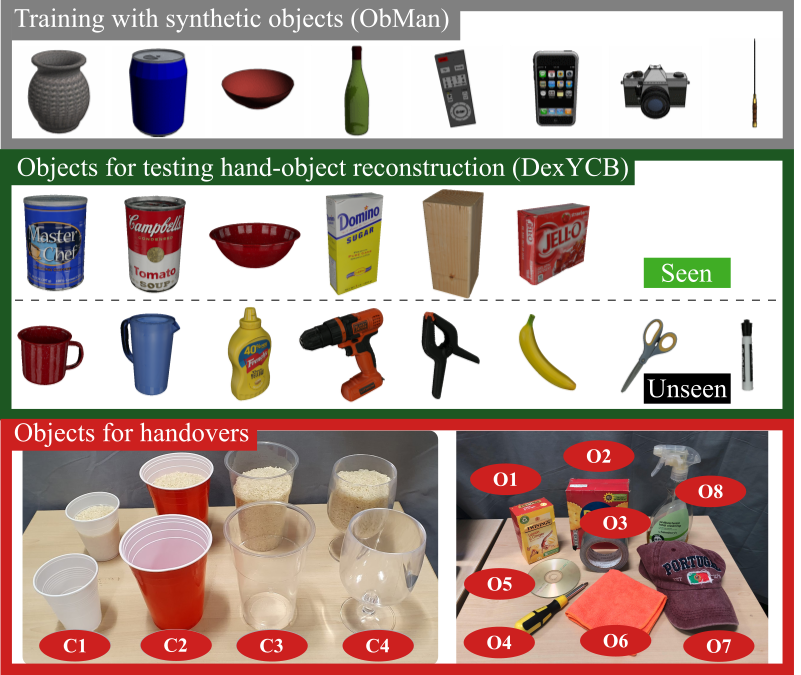}
    \caption{Synthetic object types used for training StereoHO (top row), and objects for testing hand-object reconstruction (middle row). For handovers (bottom row): 3 cups and 1 glass, empty or filled with rice, from a benchmarking protocol~\cite{sanchez2020benchmark} (left) and 8 household objects (right) to assess generalisation. 
    }
    \label{fig:objects_corsmal}
    \vspace{-10pt}
\end{figure}

We follow a benchmarking protocol~\cite{sanchez2020benchmark} that includes a setup of the robotic and perception systems and 288 handover configurations to execute. The setup  includes a 6-DoF Universal Robots UR5 robotic arm, equipped with a Robotiq 2F-85 2-finger gripper for grasping the object, and two Intel Realsense D435i cameras mounted on tripods on each side of the robotic arm pointing at the workspace of the robot which is a table with the delivery target marked.
The protocol configurations are executed by 4 participants who hand over 3 cups and 1 plastic wine glass, either empty or filled with rice, using 3 different grasp types at 3 handover locations (72 configurations for each participant)\footnote{Approval for this research was obtained from the Queen Mary Ethics of Research Committee Ref:QMERC20.525 on 19/10/2021. Consent from each human subject was obtained prior to the experiments.}. The protocol instructs to initially place the object on the table at the centre of the delivery target, and then a person picks up and hands the object over to the robot. We assume the object to be already held by the person for our pipeline and therefore we asked participants to hold the object behind a starting line indicated by tape on the table. The robot starts moving towards the object once the human moves the object beyond the starting line. To assess generalisation to different novel objects outside of the benchmark, each participant executes 24 configurations for 8 household objects at the 3 different handover locations using 1 grasp type (a total of 96 configurations). These additional objects vary in size, topology, and appearance (see Fig.~\ref{fig:objects_corsmal}).

\input{visual}

\subsection{Performance measures}

We use Chamfer distance to evaluate hand and object reconstructions. The Chamfer distance between the ground-truth and predicted point sets $S$ and $\hat{S}$ is
\begin{equation}
    d_{CD}(S,\hat{S}) = \sum_{x\in S}\min_{y\in \hat{S}} ||x-y||^2_2 + \sum_{y\in \hat{S}}\min_{x\in S} ||x-y||^2_2 .
\end{equation}

To evaluate the performance of the human-to-robot handover methods, we use three measures from the CORSMAL benchmark~\cite{sanchez2020benchmark}, including delivery location ($\delta$), efficiency ($\gamma$), and delivered mass ($\mu$), and two other measures, including grasping success ($G$) and delivery success ($D$).

    For \textit{delivery location}, we measure the distance from the center of the base of the delivered container to a predefined delivery location on the table as $d$ in mm, 

    \begin{equation}
        \delta =
        \begin{cases} 
        1 - \frac{d}{\rho}, & \text{if } d < \rho, \\
        0, & \text{otherwise}. \\ 
        \end{cases}
    \end{equation}
    We set $\rho=500$~mm as the maximum accepted distance.
    
    We quantify the \textit{efficiency} of each handover configuration as the amount of time $t$ required from the moment the robot starts moving towards the object to the moment the object is delivered on the table,
    \begin{equation}
        \gamma =
        \begin{cases} 
        1 - \frac{\max(t,\eta) - \eta}{\tau - \eta}, & \text{if } t < \tau, \\
        0, & \text{otherwise}. \\ 
        \end{cases}
    \end{equation}
    The less time required, the more efficient the method is, and the higher the efficiency score $\gamma$. The score is scaled based on the expected minimum time required $\eta=1$~s and the maximum time allowed $\tau=15$~s.
    
    For the \textit{delivered mass}, when the container is filled with rice, the robot is expected to deliver the container to the delivery location without spilling the content. We quantify the performance of the delivery based on the difference between the total mass $m$ of the filled container delivered and the total mass $\hat{m}$ of the filled container before the handover,
    \begin{equation}
        \mu =
        \begin{cases} 
        1 - \frac{|m - \hat{m}|}{\hat{m}}, & \text{if } |m - \hat{m}| < \hat{m}, \\
        0, & \text{otherwise}. \\ 
        \end{cases}
    \end{equation}
    
    We measure \textit{grasping success} as the percentage of handover configurations where the robot grasps and holds the object without dropping it after retracting.
    
    We quantify \textit{delivery success} of each configuration as the successful placement of the container on the table within the maximum delivery distance $\rho$ with no content spilled:
    \begin{equation}
        D =
        \begin{cases} 
        1, & \text{if } d < \rho \text{ and } |m - \hat{m}| = 0, \\
        0, & \text{otherwise}. \\ 
        \end{cases}
    \end{equation}
Note that since the additional objects are not containers, we do not report on the delivered mass $\mu$ and we remove the condition $|m-\hat{m}| = 0$ from the delivery success $D$.

\subsection{Hand-object reconstruction from RGB}

Since our reconstruction method combines two individual single-view predictions to form the stereo prediction, we evaluate our method in both single-view and stereo settings. We compare StereoHO with IHOI~\cite{ye2022s} in the single-view setting and with SVHO~\cite{pang2024sparse} in the stereo setting. IHOI is an SDF-based method that predicts the shape of the object conditioned on the pose of the hand, and is trained end-to-end. 
SVHO is a sparse multi-view method that reconstructs hand-object shape and can be easily adapted to use only two views. For a fair comparison, we train our method StereoHO, IHOI, and SVHO, only on the synthetic data of ObMan, and evaluate the methods using DexYCB~\cite{chao2021dexycb}. We select the two views closest to the stereo views of our setup and compare the reconstruction quality (see Fig.~\ref{fig:visual}). For the single-view setting, we aggregate results from both the left and right view. We separate the testing set into seen and unseen categories based on the similarity in 3D shape between the training and testing objects (see Fig.~\ref{fig:objects_corsmal}).

\input{ho_recon_bar_split}
\input{handover_general}

In the single-view setting (see Fig.~\ref{fig:ho_recon_bar}), our method outperforms IHOI on object reconstruction by 9.71~cm\textsuperscript{2} and 12.52~cm\textsuperscript{2}, while slightly underperforming on hand reconstruction by 1.10~cm\textsuperscript{2} and 1.21~cm\textsuperscript{2} in the seen and unseen categories, respectively. In the stereo setting, our method outperforms SVHO on object reconstruction by 35.43 cm\textsuperscript{2} and 32.27 cm\textsuperscript{2}, while performing slightly better on the hand reconstruction by 0.51~cm\textsuperscript{2} and 0.13~cm\textsuperscript{2} in the seen and unseen categories, respectively.
Compared with a version of our method without segmentation masks as input, our method improves the object reconstruction while sacrificing a small amount of performance in hand reconstruction. This also results in less noisy reconstructions when comparing with IHOI in the single-view setting (see Fig.~\ref{fig:visual}).

\subsection{Human-to-robot handovers}

We compare StereoHO with a depth-based baseline (DB),  ClearGrasp~\cite{sajjan2020clear}, and the CORSMAL baseline (CB)~\cite{pang2021towards}. DB uses the pointcloud observed from the left camera. ClearGrasp is a depth completion method to handle transparent objects using depth-based sensing. For fairness, DB and ClearGrasp are integrated in the same handover pipeline.
CB~\cite{pang2021towards} is a stereo-based pipeline specifically designed for handovers of rotationally symmetric container objects. Unlike our pipeline, CB assumes the object is initially placed on the table, following the original protocol~\cite{sanchez2020benchmark}, and is upright throughout the entire handover process. The object is detected using Mask R-CNN~\cite{He2017ICCV_MaskRCNN}, pretrained on COCO~\cite{Lin2018ECCV_COCO}, and the object shape is estimated with a multi-view iterative fitting algorithm given a cylindrical prior~\cite{xompero2020multi}. For each frame, hand keypoints are estimated using OpenPose~\cite{simon2017hand} and the 3D object centroid is triangulated from the 2D centroids of the object masks tracked with SiamMask~\cite{Wang2019CVPR_SiamMask}. Grasp points are generated along the height of the container and those ones close to the hand keypoints are removed. The robot starts moving once the object is lifted off the table. 

\input{recon_compare}
Tab.~\ref{tab:corsmalbenchmark} compares the handover performance of the methods. For the containers from the CORSMAL benchmark~\cite{sanchez2020benchmark}, both CB and our method achieve over 75\% grasping success and 65\% delivery success. This shows that assuming a cylindrical prior and an upright pose throughout the task is reasonable when handing over containers~\cite{pang2021towards}. Note that objects that are small relative to the hand (e.g., the small white cup) are less common in the training data of StereoHO and hence are more challenging to reconstruct and grasp. Moreover, the added components compared to CB, such as the reconstruction and grasp estimation modules, contribute to a higher computational load, lowering the efficiency score.
When handing over household objects, our method achieves 75\% grasping success on the opaque objects and 91\% grasping success on the transparent containers, significantly outperforming DB and ClearGrasp. Comparing the hand and object shape reconstructed by each method (see Fig.~\ref{fig:recon_compare}), DB is affected by the transparency of the object and noisy depth measurements, leaving holes in the reconstructed shape. ClearGrasp fills in the missing depth value. However, the estimated value is often inaccurate and noisy. StereoHO reconstructs a complete shape that helps estimate successful grasp candidates for the handover.

\begin{figure}[t!]
    \centering
    \includegraphics[width=1.0\columnwidth,trim={0.25cm 0 0.3cm 0},clip]{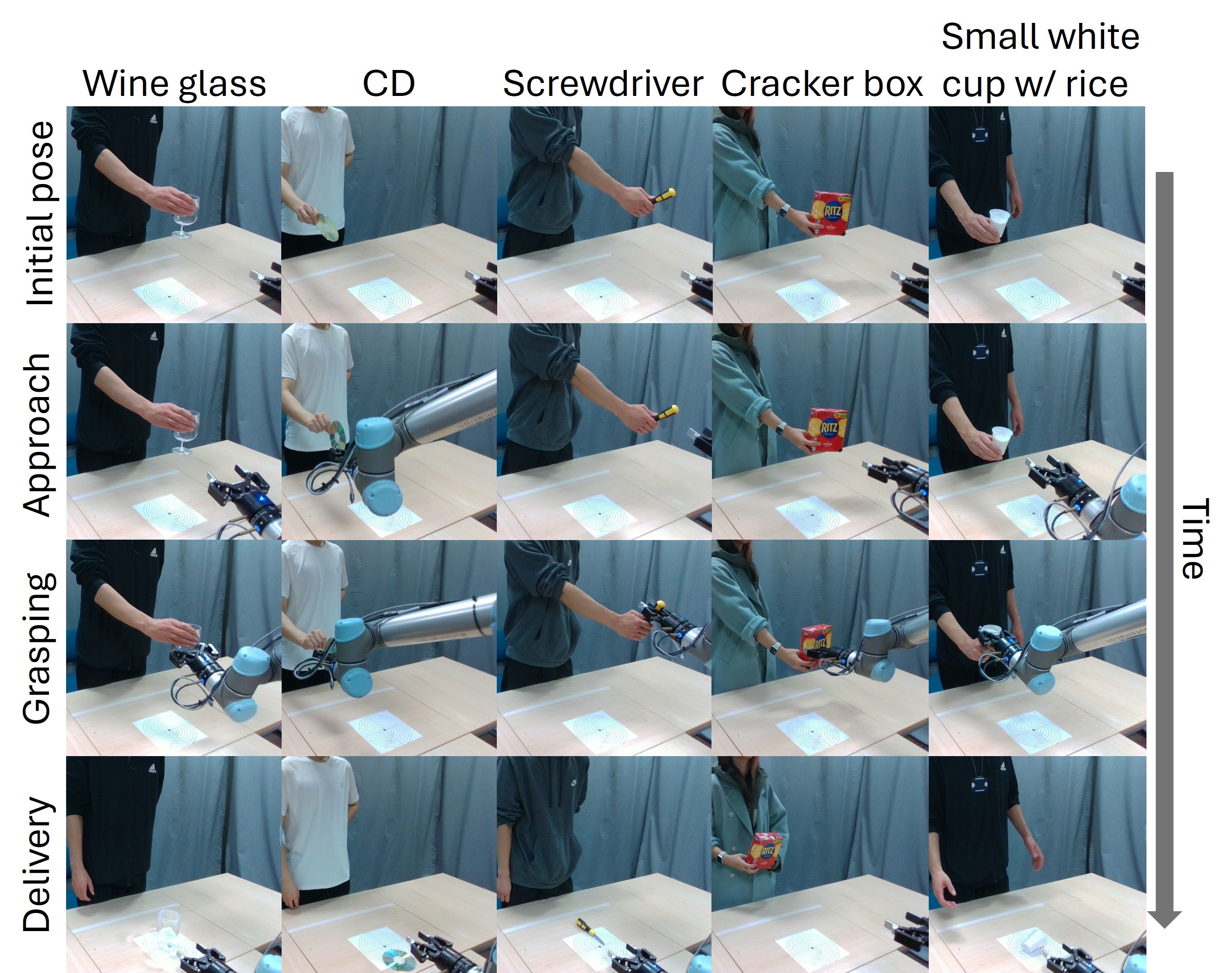}\\
    \caption{Examples of successful human-to-robot handovers of household objects using StereoHO, grasping failures, and spilling of the content.}
    \label{fig:handover_visual2}
    \vspace{-10pt}
\end{figure}
Overall, our method outperforms depth-based methods when handing over general household objects (see Fig.~\ref{fig:handover_visual2}). For the handover of containers, our method performs on par with CB, which is specifically designed for the handover of containers.
Specifically, our method successfully receives objects with challenging properties, such as thin shape (screwdriver), deformable shape (cloth), and reflective material (CD). On the contrary, CB fails to generalise to non-container objects, and depth-based methods (DB and ClearGrasp) fail to receive transparent objects. Failures occur for StereoHO when the reconstruction of the object shape is inaccurate. For example, the robot often try to grasp the cracker box from the wider side (see Fig.~\ref{fig:handover_visual2}), which is larger than the gripper width. Failure can also occur when the robot loses track of the hand due to occlusions by the object or the robot arm itself.

\section{Conclusion}
\label{sec:conclusion}

In this paper, we proposed a joint hand-object reconstruction method to enable human-to-robot handovers using wide baseline stereo RGB cameras. Our reconstruction method outperforms IHOI and SVHO in both single-view and stereo settings. Handover experiments showed that our method enables a robot to receive empty and filled containers, and can generalize to other household objects. Future work includes improving the speed of handovers and the quality of hand-object reconstructions.

\balance

{\small
\bibliographystyle{IEEEtran}
\bibliography{refs}
}

\end{document}

%% file: robot_setup.tex
\begin{figure}[t!]
    \centering
    \includegraphics[width=\columnwidth,trim={2cm 2cm 0 0},clip]{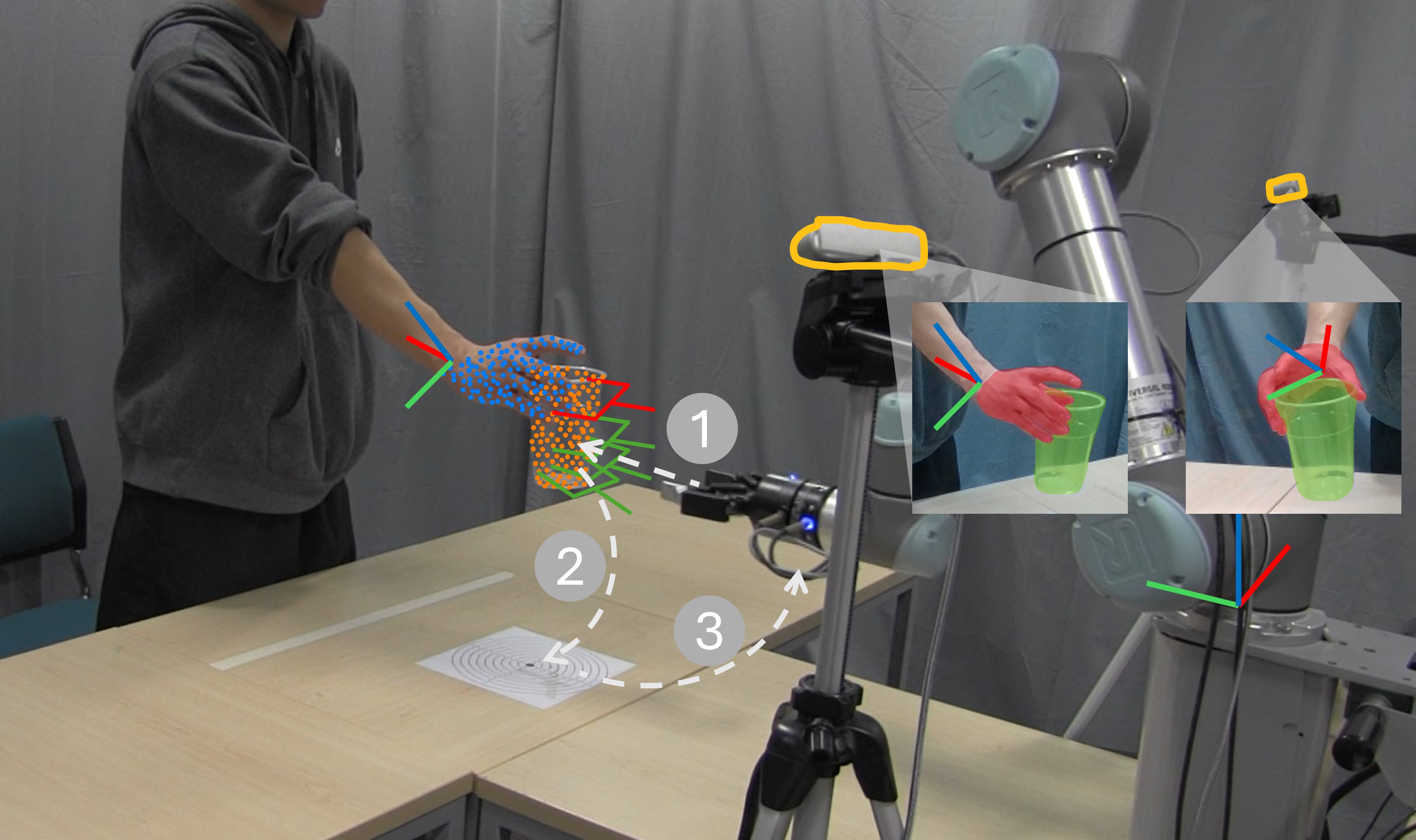}
    \caption{We reconstruct the hand-object pointcloud from stereo RGB input for human-to-robot handover. (1) A safe grasp is selected for the handover and the robot moves in to grasp the object. (2) The object is delivered to a target location. (3) The robot returns to its home position.}
    \label{fig:robot_setup}
    \vspace{-10pt}
\end{figure}

%% file: overview_recon.tex
\begin{figure*}[t!]
    \centering
    \includegraphics[width=\linewidth,trim={0cm 0 0 0},clip]{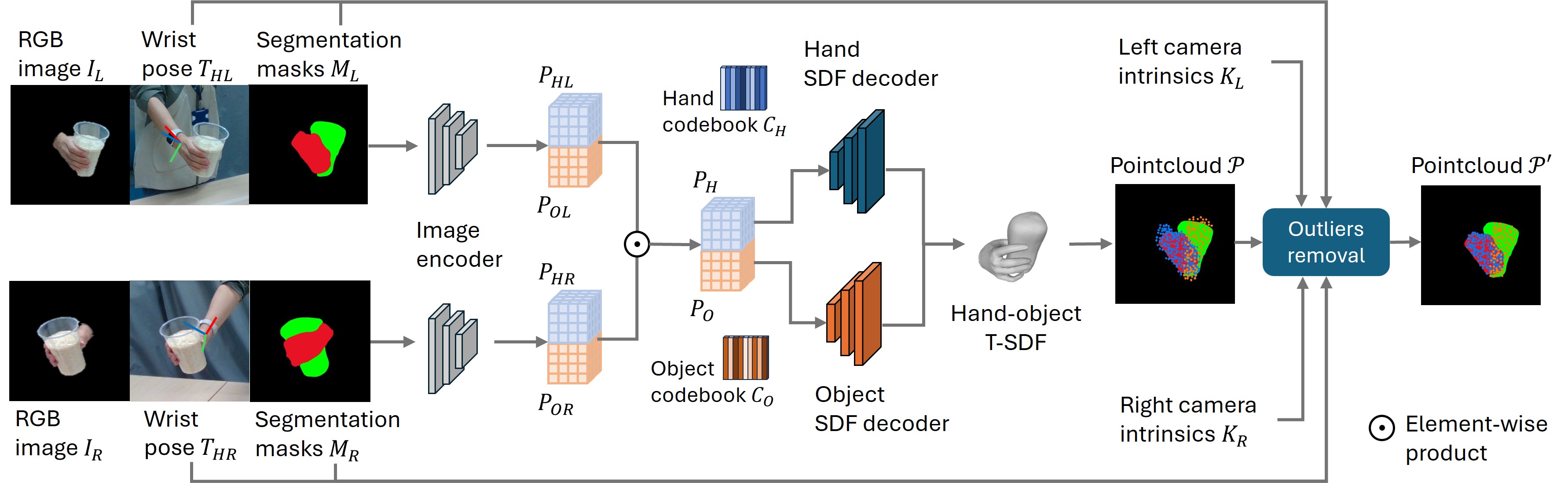}\\
    \caption{Our proposed method, StereoHO, for hand-object reconstruction with two cropped images from a wide-baseline stereo camera. StereoHO predicts the probability distributions, $P_{L} = [P_{HL}, P_{OL}]$ and $P_{R} = [P_{HR}, P_{OR}]$, over the shape codebooks for each view and combined them into a coherent probability distribution $P = [P_{H}, P_{O}]$. The trained SDF decoder transforms $P$ into the hand-object T-SDF. Sampled surface points from the T-SDF (pointcloud $\mathcal{P}$) are projected into each view using the predicted camera projection parameters and outliers are removed by using the segmentation masks to obtain the final pointcloud $\mathcal{P}'$. KEYS -- $K$:~intrinsics calibration parameters, $L$:~left view, $R$:~right view; $H$:~hand, $O$:~object.}
    \label{fig:overview_recon}
\end{figure*}

%% file: overview_robot.tex
\begin{figure*}[t!]
    \centering
    \includegraphics[width=1.0\linewidth,trim={0cm 0 0 0},clip]{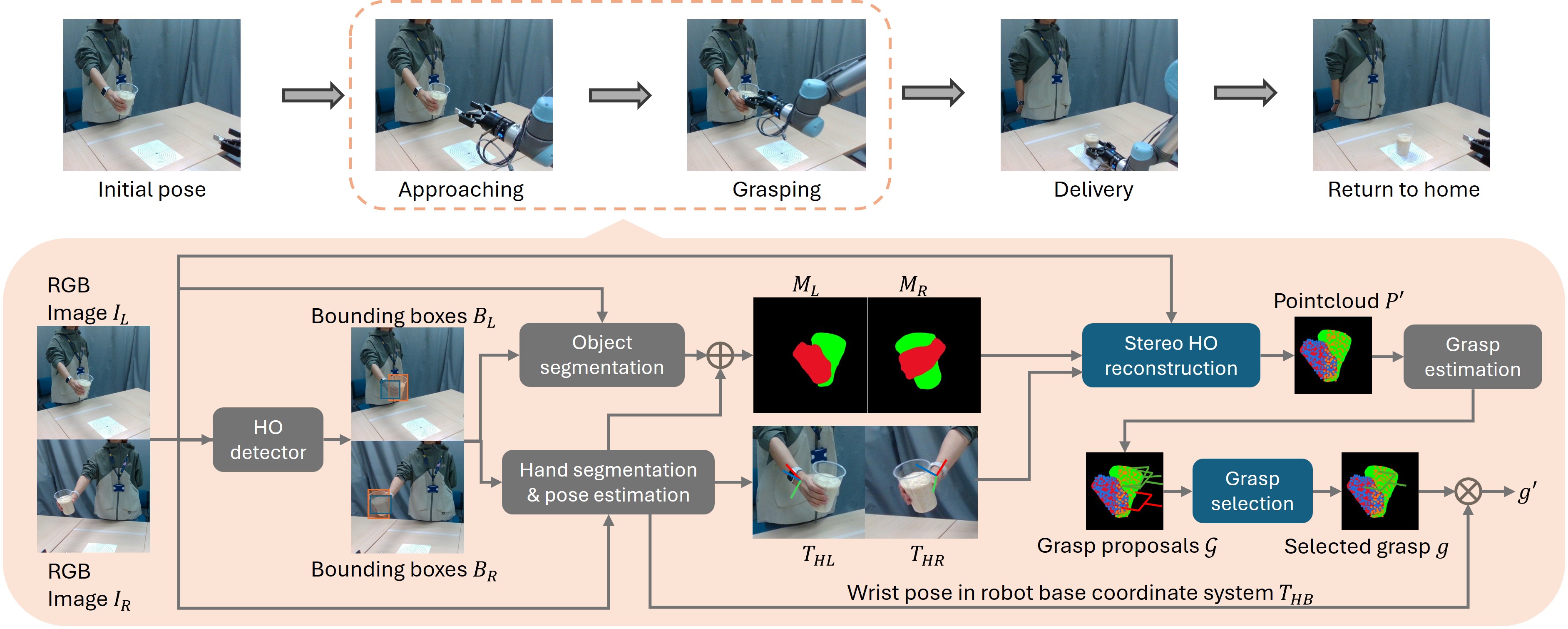}\\
    \caption{Our proposed pipeline for human-to-robot handover. For each frame, hand-object detection extracts bounding boxes $B$ to crop the input images around the hand and then used to segment hand-object masks $M$ and estimate wrist poses $T_{H}$. Our StereoHO uses these outputs along with the image crops to reconstruct the hand-object shape $\mathcal{P}'$. We estimate the grasp $g$ on the reconstructed shape and transform it into the robot coordinate system using the wrist poses $T_{HB}$.}
    \label{fig:overview_robot}
\end{figure*}

%% file: visual.tex
\begin{figure*}[t!]
    \centering
    \includegraphics[width=0.49\linewidth]{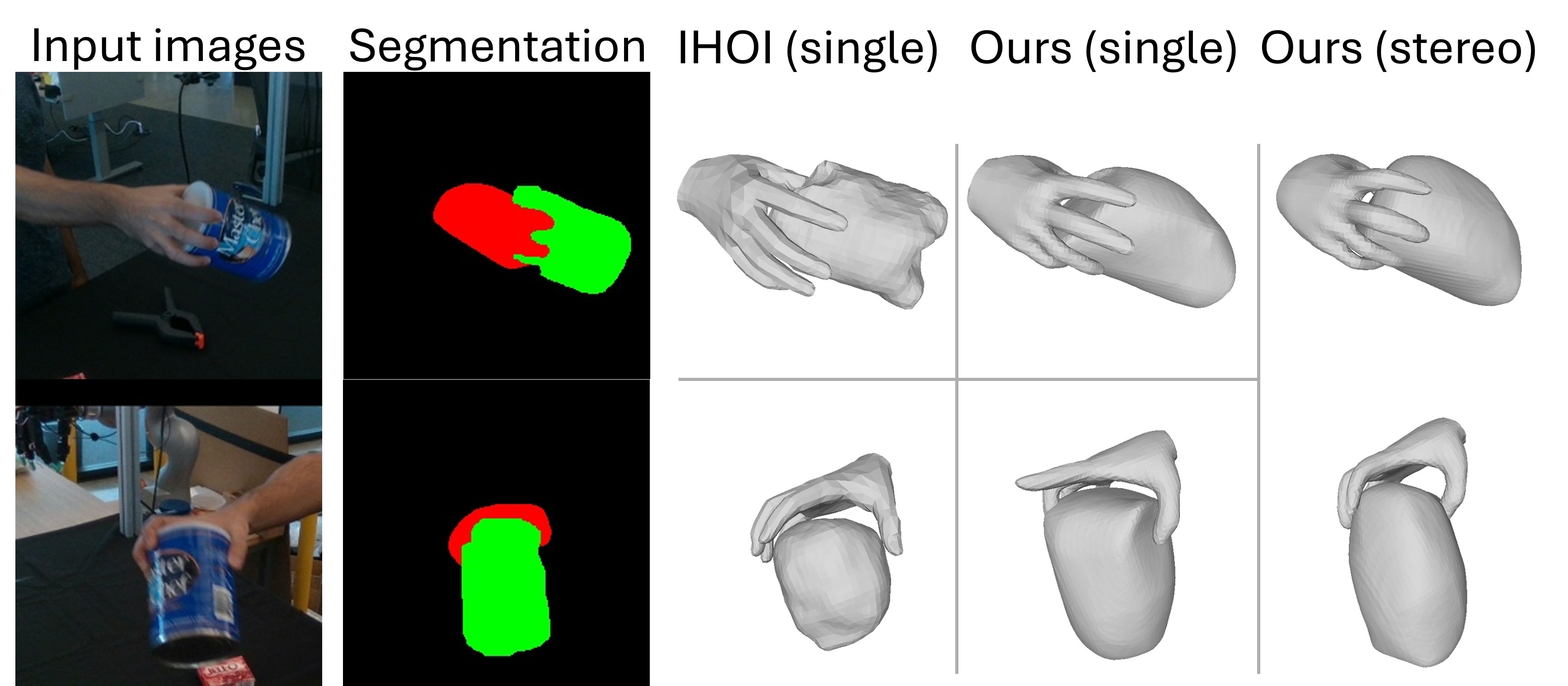}
    \includegraphics[width=0.49\linewidth]{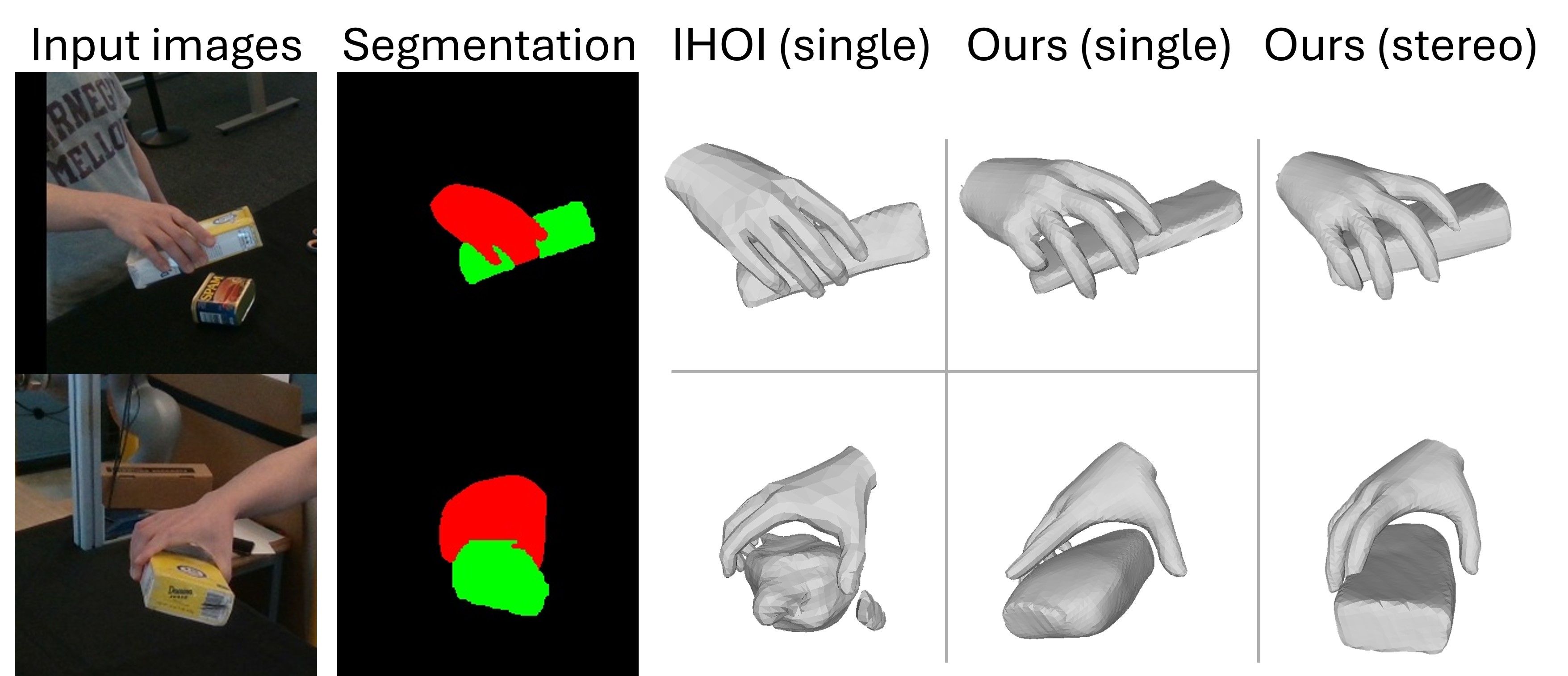}
    \caption{Comparing single-view and stereo hand object reconstructions on DexYCB~\cite{chao2021dexycb}. For single-view, each reconstruction corresponds to the image on the same row. For stereo, the same reconstruction is shown from two viewpoints. Our method yields less noisy reconstructions and improves both the hand and object shape by combining the single-view predictions.
    }
    \label{fig:visual}
    \vspace{-10pt}
\end{figure*}

%% file: ho_recon_bar_split.tex
\begin{figure}
    \centering
    \begin{tikzpicture}
        \begin{axis}[
            width=0.55\linewidth,
            height=0.4\columnwidth,
            xlabel={CD [cm$^2$] (Object)},
            ylabel={CD [cm$^2$] (Hand)},
            ymin=1, ymax=3.2,
            xmin=10, xmax=70,
            xtick={10,20,30,40,50,60,70},
            xlabel near ticks,
            ylabel near ticks,
            label style={font=\footnotesize},
            tick label style={font=\scriptsize},
            title={{\bf Single seen}},
            title style={font=\footnotesize, yshift=-6pt},
            scatter/classes={
            aa={mark=*,dc1},
            ba={mark=*,dc2},
            ca={mark=*,dc5},
            da={mark=*,dc4},
            ab={mark=square*,dc1},
            bb={mark=square*,dc2},
            cb={mark=square*,dc3},
            db={mark=square*,dc4}
            }
        ]
        \addplot+[dashed, color=gray,mark=none] coordinates{(20,0) (20,100)};
        \addplot+[dashed, color=gray,mark=none] coordinates{(30,0) (30,100)};
        \addplot+[dashed, color=gray,mark=none] coordinates{(40,0) (40,100)};
        \addplot+[dashed, color=gray,mark=none] coordinates{(50,0) (50,100)};
        \addplot+[dashed, color=gray,mark=none] coordinates{(60,0) (60,100)};
        \addplot+[scatter, only marks, scatter src=explicit symbolic] table[x=obj,y=hand, meta=label]{
        obj hand    label
        62.62    2.98    aa
        38.77    1.46    ba
        40.28    2.49    ca
        29.06    2.56    da
        };
        \end{axis}
    \end{tikzpicture}
    \begin{tikzpicture}
        \begin{axis}[
            width=0.55\linewidth,
            height=0.4\columnwidth,
            xlabel={CD [cm$^2$] (Object)},
            ymin=1, ymax=3.2,
            xmin=10, xmax=70,
            xtick={10,20,30,40,50,60,70},
            yticklabels={},
            xlabel near ticks,
            ylabel near ticks,
            label style={font=\footnotesize},
            tick label style={font=\scriptsize},
            title={{\bf Single unseen}},
            title style={font=\footnotesize,yshift=-6pt},
            scatter/classes={
            aa={mark=*,dc1},
            ba={mark=*,dc2},
            ca={mark=*,dc5},
            da={mark=*,dc4}
            }
        ]
        \addplot+[dashed, color=gray,mark=none] coordinates{(20,0) (20,100)};
        \addplot+[dashed, color=gray,mark=none] coordinates{(30,0) (30,100)};
        \addplot+[dashed, color=gray,mark=none] coordinates{(40,0) (40,100)};
        \addplot+[dashed, color=gray,mark=none] coordinates{(50,0) (50,100)};
        \addplot+[dashed, color=gray,mark=none] coordinates{(60,0) (60,100)};
        \addplot+[scatter, only marks, scatter src=explicit symbolic] table[x=obj,y=hand, meta=label]{
        obj hand    label
        69.55    2.44    aa
        55.98    1.13    ba
        47.25    2.24    ca
        43.46    2.34    da
        };
        \end{axis}
    \end{tikzpicture}
    \begin{tikzpicture}
        \begin{axis}[
            width=0.55\linewidth,
            height=0.4\columnwidth,
            xlabel={CD [cm$^2$] (Object)},
            ylabel={CD [cm$^2$] (Hand)},
            ymin=1, ymax=3.2,
            xmin=10, xmax=70,
            xtick={10,20,30,40,50,60,70},
            xlabel near ticks,
            ylabel near ticks,
            label style={font=\footnotesize},
            tick label style={font=\scriptsize},
            title={{\bf Stereo seen}},
            title style={font=\footnotesize, yshift=-6pt},
            scatter/classes={
            aa={mark=*,dc1},
            ba={mark=*,dc2},
            ca={mark=*,dc5},
            da={mark=*,dc4}
            }
        ]
        \addplot+[dashed, color=gray,mark=none] coordinates{(20,0) (20,100)};
        \addplot+[dashed, color=gray,mark=none] coordinates{(30,0) (30,100)};
        \addplot+[dashed, color=gray,mark=none] coordinates{(40,0) (40,100)};
        \addplot+[dashed, color=gray,mark=none] coordinates{(50,0) (50,100)};
        \addplot+[dashed, color=gray,mark=none] coordinates{(60,0) (60,100)};
        \addplot+[scatter, only marks, scatter src=explicit symbolic] table[x=obj,y=hand, meta=label]{
        obj hand    label
        53.96    2.86    aa
        0    0    ba
        25.09    2.24    ca
        18.53    2.35    da
        };
        \end{axis}
    \end{tikzpicture}
    \begin{tikzpicture}
        \begin{axis}[
            width=0.55\linewidth,
            height=0.4\columnwidth,
            xlabel={CD [cm$^2$] (Object)},
            ymin=1, ymax=3.2,
            xmin=10, xmax=70,
            xtick={10,20,30,40,50,60,70},
            xlabel near ticks,
            ylabel near ticks,
            yticklabels={},
            label style={font=\footnotesize},
            tick label style={font=\scriptsize},
            title={{\bf Stereo unseen}},
            title style={font=\footnotesize, yshift=-6pt},
            scatter/classes={
            aa={mark=*,dc1},
            ba={mark=*,dc2},
            ca={mark=*,dc5},
            da={mark=*,dc4}
            }
        ]
        \addplot+[dashed, color=gray,mark=none] coordinates{(20,0) (20,100)};
        \addplot+[dashed, color=gray,mark=none] coordinates{(30,0) (30,100)};
        \addplot+[dashed, color=gray,mark=none] coordinates{(40,0) (40,100)};
        \addplot+[dashed, color=gray,mark=none] coordinates{(50,0) (50,100)};
        \addplot+[dashed, color=gray,mark=none] coordinates{(60,0) (60,100)};
        \addplot+[scatter, only marks, scatter src=explicit symbolic] table[x=obj,y=hand, meta=label]{
        obj hand    label
        61.83    2.3    aa
        0    0    ba
        33.90    2.04    ca
        29.56    2.17    da
        };
        \end{axis}
    \end{tikzpicture}
    \caption{Single-view and stereo hand-object reconstruction errors on DexYCB~\cite{chao2021dexycb}.
    Best results on the bottom-left corner.
    Legend -- 
    CD:~Chamfer distance,
    {\protect\raisebox{1pt}{\protect\tikz \protect\draw[dc1,fill=dc1]    (1,1) circle (0.5ex);}}~SVHO~\cite{pang2024sparse},
    {\protect\raisebox{1pt}{\protect\tikz \protect\draw[dc2,fill=dc2] (1,1) circle (0.5ex);}}~IHOI~\cite{ye2022s},
    {\protect\raisebox{1pt}{\protect\tikz \protect\draw[dc5,fill=dc5] (1,1) circle (0.5ex);}}~Ours (no segmentation mask),
    {\protect\raisebox{1pt}{\protect\tikz \protect\draw[dc4,fill=dc4] (1,1) circle (0.5ex);}}~Ours.
    IHOI (single-view reconstruction method) is not evaluated in stereo setting.}
    \label{fig:ho_recon_bar}
    \vspace{-10pt}
\end{figure}

%% file: handover_general.tex
\begin{table*}[t!]
    \centering
    \footnotesize
    \setlength\tabcolsep{2.4pt}
    \caption{Comparison of performance results between hand-object reconstruction methods integrated in a system for human-to-robot handovers of unknown and diverse household objects with varying physical properties (shape, mass, transparency).
    }
    \vspace{-5pt}
    \begin{tabular}{cccccccc|ccccc|ccccccccc|ccc}
    \toprule
    \textbf{Score} & \textbf{Method} & \textbf{DoF} & \textbf{RGB} & \textbf{D} & \textbf{DC} & \textbf{SC} & \textbf{HH} & \multicolumn{5}{c}{\textbf{CORSMAL containers}} & \multicolumn{12}{c}{\textbf{Household objects}} \\
    \cmidrule(lr){9-13}\cmidrule(lr){14-25}
    & & & & & & & &  Avg & \cellcolor{o}C1 & \cellcolor{o}C2 & \cellcolor{m}C3 & \cellcolor{m}C4 & Avg & \cellcolor{o}O1 & \cellcolor{o}O2 & \cellcolor{o}O3 & \cellcolor{o}O4 & \cellcolor{o}O5 & \cellcolor{o}O6 & \cellcolor{o}O7 & \cellcolor{o}O8 & Avg & \cellcolor{t}C3 & \cellcolor{t}C4\\
    \midrule
    \multirow{4}{*}{$\delta$}
    & CB~\cite{pang2021towards} & 3 & \bbox & \wbox & \wbox & \bbox & \wbox & 0.45 & \textbf{0.52} & 0.58 & 0.33 & 0.38 & - & - & - & - & - & - & - & - & - & 0.31 & 0.05 & \textbf{0.57} \\
    & DB & 6 & \bbox & \bbox & \wbox & \wbox & \bbox & - & - & - & - & - & 0.06 & 0.05 & 0.00 & 0.05 & 0.06 & 0.11 & 0.00 & 0.00 & 0.20 & 0.01 & 0.03 & 0.00 \\
    & ClearGrasp~\cite{sajjan2020clear} & 6 & \bbox & \bbox & \bbox & \wbox & \bbox & - & - & - & - & - & 0.01 & 0.00 & 0.00 & 0.00 & 0.00 & 0.00 & 0.05 & 0.00 & 0.08 & 0.00 & 0.00 & 0.00 \\
    \rowcolor{mylightgray} \cellcolor{white} & StereoHO & 6 & \bbox & \wbox & \wbox & \bbox & \bbox & \textbf{0.48} & 0.26 & \textbf{0.65} & \textbf{0.59} & \textbf{0.42} & \textbf{0.18} & \textbf{0.18} & \textbf{0.11} & \textbf{0.06} & \textbf{0.25} & \textbf{0.31} & \textbf{0.09} & \textbf{0.18} & \textbf{0.23} & \textbf{0.55} & \textbf{0.65} & 0.45 \\    
    \midrule
    \multirow{4}{*}{$\gamma$}
    & CB~\cite{pang2021towards} & 3 & \bbox & \wbox & \wbox & \bbox & \wbox & \textbf{0.29} & \textbf{0.31} & \textbf{0.35} & \textbf{0.26} & \textbf{0.25} & - & - & - & - & - & - & - & - & - & \textbf{0.20} & \textbf{0.16} & \textbf{0.24} \\
    & DB & 6 & \bbox & \bbox & \wbox & \wbox & \bbox & - & - & - & - & - & \textbf{0.29} & \textbf{0.34} & \textbf{0.27} & \textbf{0.30} & \textbf{0.27} & \textbf{0.37} & 0.27 & 0.10 & \textbf{0.38} & 0.13 & 0.10 & 0.16 \\
    & ClearGrasp~\cite{sajjan2020clear} & 6 & \bbox & \bbox & \bbox & \wbox & \bbox & - & - & - & - & - & 0.14 & 0.15 & 0.05 & 0.06 & 0.14 & 0.15 & 0.28 & 0.14 & 0.16 & 0.09 & 0.10 & 0.08 \\
    \rowcolor{mylightgray} \cellcolor{white} & StereoHO & 6 & \bbox & \wbox & \wbox & \bbox & \bbox & 0.10 & 0.12 & 0.09 & 0.10 & 0.09 & 0.21 & 0.09 & 0.07 & 0.07 & 0.14 & 0.13 & \textbf{0.39} & \textbf{0.44} & 0.35 & 0.09 & 0.09 & 0.09 \\
    \midrule
    \multirow{1}{*}{$\mu$}
    & CB~\cite{pang2021towards} & 3 & \bbox & \wbox & \wbox & \bbox & \wbox & \textbf{0.78} & \textbf{0.77} & 0.81 & 0.77 & \textbf{0.78} & - & - & - & - & - & - & - & - & - & - & - & - \\
    \rowcolor{mylightgray} \cellcolor{white} & StereoHO & 6 & \bbox & \wbox & \wbox & \bbox & \bbox & 0.74 & 0.53 & \textbf{0.87} & \textbf{0.84} & 0.73 & - & - & - & - & - & - & - & - & - & - & - & - \\
    \midrule
    \multirow{4}{*}{$G$}
    & CB~\cite{pang2021towards} & 3 & \bbox & \wbox & \wbox & \bbox & \wbox & \textbf{0.79} & \textbf{0.77} & 0.81 & 0.79 & \textbf{0.80} & - & - & - & - & - & - & - & - & - & 0.83 & 0.66 & \textbf{1.00} \\
    & DB & 6 & \bbox & \bbox & \wbox & \wbox & \bbox & - & - & - & - & - & 0.41 & 0.16 & 0.16 & 0.50 & 0.50 & 0.58 & 0.50 & 0.08 & \textbf{0.83} & 0.08 & 0.08 & 0.08 \\
    & ClearGrasp~\cite{sajjan2020clear} & 6 & \bbox & \bbox & \bbox & \wbox & \bbox & - & - & - & - & - & 0.15 & 0.00 & 0.00 & 0.08 & 0.25 & 0.16 & 0.41 & 0.00 & 0.33 & 0.20 & 0.33 & 0.08 \\
    \rowcolor{mylightgray} \cellcolor{white} & StereoHO & 6 & \bbox & \wbox & \wbox & \bbox & \bbox & 0.75 & 0.55 & \textbf{0.88} & \textbf{0.84} & 0.73 & \textbf{0.75} & \textbf{0.50} & \textbf{0.66} & \textbf{0.75} & \textbf{0.75} & \textbf{0.83} & \textbf{0.75} & \textbf{1.00} & 0.75 & \textbf{0.91} & \textbf{1.00} & 0.83 \\    
    \midrule
    \multirow{4}{*}{$D$}
    & CB~\cite{pang2021towards} & 3 & \bbox & \wbox & \wbox & \bbox & \wbox & \textbf{0.67} & \textbf{0.73} & 0.77 & 0.50 & \textbf{0.68} & - & - & - & - & - & - & - & - & - & 0.50 & 0.08 & \textbf{0.91} \\
    & DB & 6 & \bbox & \bbox & \wbox & \wbox & \bbox & - & - & - & - & - & 0.20 & 0.08 & 0.16 & 0.33 & 0.25 & 0.25 & 0.00 & 0.00 & \textbf{0.58} & 0.04 & 0.08 & 0.00 \\
    & ClearGrasp~\cite{sajjan2020clear} & 6 & \bbox & \bbox & \bbox & \wbox & \bbox & - & - & - & - & - & 0.05 & 0.00 & 0.00 & 0.00 & 0.00 & 0.00 & \textbf{0.16} & 0.00 & 0.25 & 0.00 & 0.00 & 0.00 \\
    \rowcolor{mylightgray} \cellcolor{white} & StereoHO & 6 & \bbox & \wbox & \wbox & \bbox & \bbox & 0.66 & 0.36 & \textbf{0.86} & \textbf{0.84} & 0.59 & \textbf{0.50} & \textbf{0.41} & \textbf{0.58} & \textbf{0.50} & \textbf{0.58} & \textbf{0.83} & 0.08 & \textbf{0.58} & 0.41 & \textbf{0.83} & \textbf{1.00} & 0.66 \\
    \bottomrule
    \addlinespace[\belowrulesep]
    \multicolumn{25}{l}{\parbox{0.98\linewidth}{\scriptsize{Results of our proposed method highlighted in \colorbox{mylightgray}{\strut gray}. Objects highlighted based on their transparency (\colorbox{t}{\strut transparent}, \colorbox{o}{\strut opaque}, \colorbox{m}{\strut mixed}). Best scores highlighted in \textbf{bold} for each column and each performance measure.
    KEY -- DoF:~degrees of freedom, RGB:~colour input, D:~depth input, DC:~depth completion, SC:~stereo camera, HH:~handheld object assumption, Avg:~average, C1:~small white cup, C2:~red cup, C3:~beer cup, C4:~wine glass, O1:~tea box, O2:~cracker box, O3:~tape, O4:~screwdriver, O5:~CD, O6:~cloth, O7:~cap, O8:~spray bottle; $\delta$:~delivery location, $\gamma$:~efficiency, $\mu$:~delivered mass, $G$:~grasping success, $D$:~delivery success, \wbox:~not used, \bbox:~used; -:~not applicable.}}}
    \end{tabular}
    \label{tab:corsmalbenchmark}
    \vspace{-10pt}
\end{table*}

%% file: recon_compare.tex
\begin{figure}[t!]
    \centering
    \includegraphics[width=0.9\columnwidth,trim={0 0.5cm 0 0},clip]{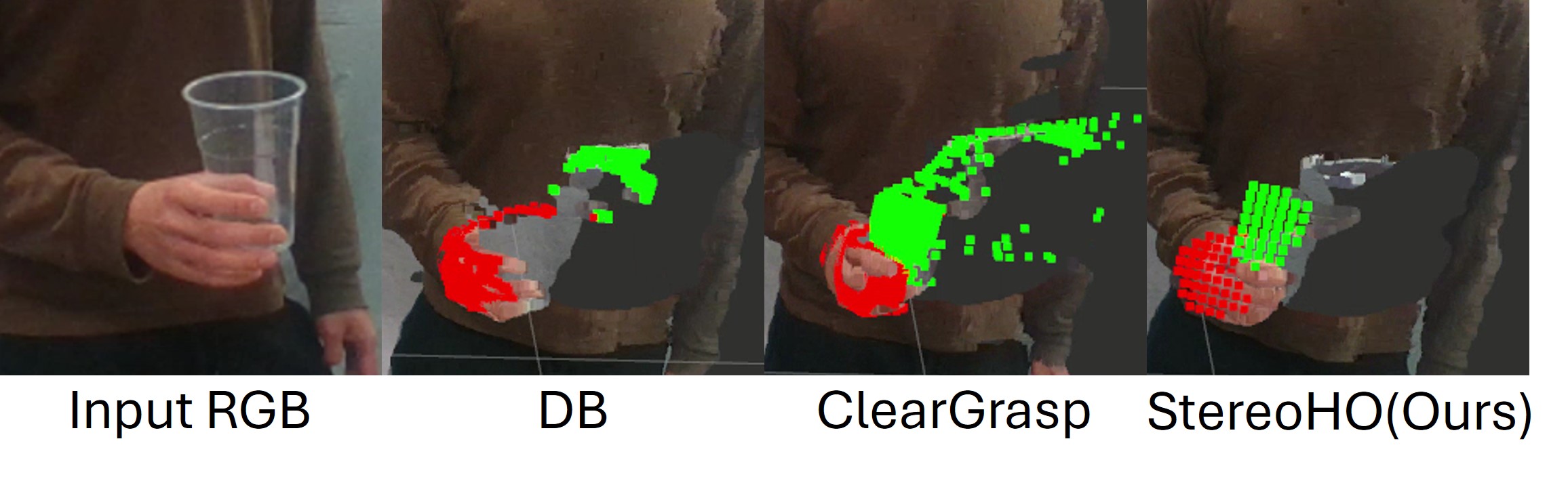}
    \caption{Comparing hand (red) and object (green) reconstructions, overlaid on the pointcloud obtained by the camera.}
    \label{fig:recon_compare}
    \vspace{-10pt}
\end{figure}